\documentclass[conference]{IEEEtran}
\IEEEoverridecommandlockouts

\usepackage{cite}
\usepackage{amsmath,amssymb,amsfonts}
\usepackage{algorithm}
\usepackage{algorithmic}
\usepackage{graphicx}
\usepackage{booktabs}
\usepackage{multirow}
\usepackage{placeins}
\usepackage{textcomp}
\usepackage{url}
\usepackage{xcolor}

\def\BibTeX{{\rm B\kern-.05em{\sc i\kern-.025em b}\kern-.08em
    T\kern-.1667em\lower.7ex\hbox{E}\kern-.125emX}}

\begin{document}

\title{SHAPE: Coalition-Aware Expert Pruning for Sparse Mixture-of-Experts LLMs}
\author{\IEEEauthorblockN{Yuhao Zhang, Beihang University, yuhao\_zh@buaa.edu.cn}}
\maketitle

\begin{abstract}
Sparse Mixture-of-Experts (MoE) large language models achieve strong quality with low per-token compute, yet their deployment is often limited by the memory wall: the full expert pool must remain resident to support token-dependent routing. Expert pruning is a direct remedy, but prior criteria typically score experts independently and overlook that MoE inference is inherently \emph{coalitional}, where outputs arise from routed top-$k$ expert combinations. We propose \textbf{SHAPE}, a task-driven pruning framework that explicitly models \emph{intra-layer} expert cooperation. SHAPE formulates routing traces on a small calibration set as an empirical cooperative game and assigns interaction-aware expert values via a Shapley-style attribution over observed top-$k$ coalitions, enabling the identification of experts that are essential for high-utility collaborations rather than merely frequent. To preserve MoE topology under a global pruning budget, SHAPE further introduces a \emph{quality-coverage} selection rule that retains, in each layer, the minimal expert subset covering an $\alpha$ fraction of non-negative Shapley mass, while using bisection to match a target keep rate. Experiments on three modern MoE backbones (Qwen3-30B-A3B, GPT-OSS-20B, and DeepSeek-V2-Lite) across diverse benchmarks show that SHAPE consistently improves robustness over global and layer-wise pruning variants, maintaining competitive accuracy under 20\% and 40\% expert pruning without additional training and delivering clear reductions in peak GPU memory footprint. The open-source code is available at \url{https://github.com/Alizen-1009/Shapley-Moe}.
\end{abstract}


\section{Introduction}
\label{sec:intro}

The landscape of Large Language Models (LLMs) has been fundamentally reshaped by the widespread adoption of Sparse Mixture-of-Experts (SMoE) architectures. 
Pioneering works such as GShard~\cite{lepikhin2020gshard} and Switch Transformers~\cite{fedus2021switch}, followed by recent state-of-the-art open models including gpt-oss~\cite{openai2025gptoss120bgptoss20bmodel} and DeepSeek-V3.2~\cite{deepseekai2025deepseekv32pushingfrontieropen}, have demonstrated that model capacity can be scaled to hundreds of billions or even trillions of parameters with sub-linear growth in training cost. 
By decoupling total parameter count from per-token active computation, SMoE architectures enable a favorable trade-off between model capacity and inference latency, leading to strong performance on complex reasoning and coding tasks~\cite{yang2025qwen3technicalreport}.

However, despite their favorable computational efficiency, the deployment of massive MoE models encounters a critical bottleneck: the ``Memory Wall.'' 
While conditional computation ensures low floating-point operations (FLOPs) per token, the \emph{entire} expert pool must reside in GPU memory to support dynamic, token-dependent routing decisions. 
For example, serving a 671B-parameter model such as DeepSeek-V3.2 requires a multi-node GPU cluster with terabytes of VRAM, regardless of its sparse activation pattern~\cite{deepseekai2025deepseekv32pushingfrontieropen}. 
This prohibitive memory requirement fundamentally limits the deployability of high-capacity MoE models in resource-constrained environments, motivating an urgent need for post-training compression techniques that can substantially reduce the served parameter footprint while preserving the functional behavior of routed expert subnetworks.

Among emerging compression strategies, \emph{expert pruning} has emerged as a prominent approach for reducing the parameter footprint of Mixture-of-Experts models by selectively removing experts. 
Unlike quantization~\cite{qmoe2023, hu2025moequant}, which reduces numerical precision, or expert merging~\cite{ li2025moesvd, zhou2025dernDroppingExpertsRecombiningNeurons}, which fuses expert parameters and may alter underlying functional subspaces~\cite{bai2025diep}, pruning aims to reduce model size by selecting a subset of experts to serve at inference time. 
Early pruning approaches predominantly relied on heuristic, expert-wise metrics, such as activation frequency~\cite{chen2022task} or gating magnitude~\cite{muzio2024seer}. 
More recent methods incorporate training dynamics or activation statistics to refine expert-wise saliency estimation, including router-weight shift metrics~\cite{chowdhury2024provably} and activation-aware scoring methods \cite{liang2025seap, lasby2025reap}.

Despite these methodological advancements, a fundamental modeling gap remains: existing pruning criteria predominantly adopt an \emph{expert-wise} perspective~\cite{xu2025camera, cao2025condense, zhang2025diversifyingexpertknowledgetaskagnostic}, implicitly assuming that experts contribute independently to the model output. 
This assumption conflicts with the intrinsic nature of MoE inference, which is inherently \emph{coalitional}. 
For each token, the model output is produced by a non-linear composition of a specific coalition of top-$k$ experts selected by the router ~\cite{shazeer2017outrageously}. 
Consequently, the functional utility of an expert is fundamentally conditioned on the set of co-activated experts within the routed coalition; an expert may exhibit limited standalone utility yet provide indispensable complementary functionality when paired with semantically dominant experts~\cite{su2025unveilingsuperexpertsmixtureofexperts}. 
Conversely, an expert with high routing frequency may be functionally redundant if its contribution is subsumed by other members of the coalition. 
By ignoring such intra-layer cooperative interactions, expert-wise pruning risks dismantling high-utility coalitions, resulting in disproportionate performance degradation on complex downstream tasks.

To address this modeling limitation, we introduce \textbf{SHAPE}, a Shapley-guided pruning framework that explicitly models expert selection as a cooperative process. 
Rather than relying on static, expert-wise heuristics, SHAPE formulates MoE inference on a calibration set~\cite{cobbe2021gsm8k, peng2024humanevalxl} as a cooperative game, where the routed top-$k$ experts constitute a coalition of players. We employ Shapley values~\cite{shapley1953}, the unique valuation scheme satisfying standard fairness axioms, to attribute each expert’s marginal contribution to the utility of the routed coalition. 
This interaction-aware valuation enables SHAPE to distinguish experts that are merely frequently selected from those that are cooperatively essential to high-utility coalitions. 
Furthermore, to avoid structural collapse caused by overly aggressive pruning, we introduce a \emph{Shapley Quality Coverage} mechanism that enforces layer-wise retention based on the distribution of cooperative value.
 
\begin{figure*}[t]
\centering
\includegraphics[width=0.95\textwidth]{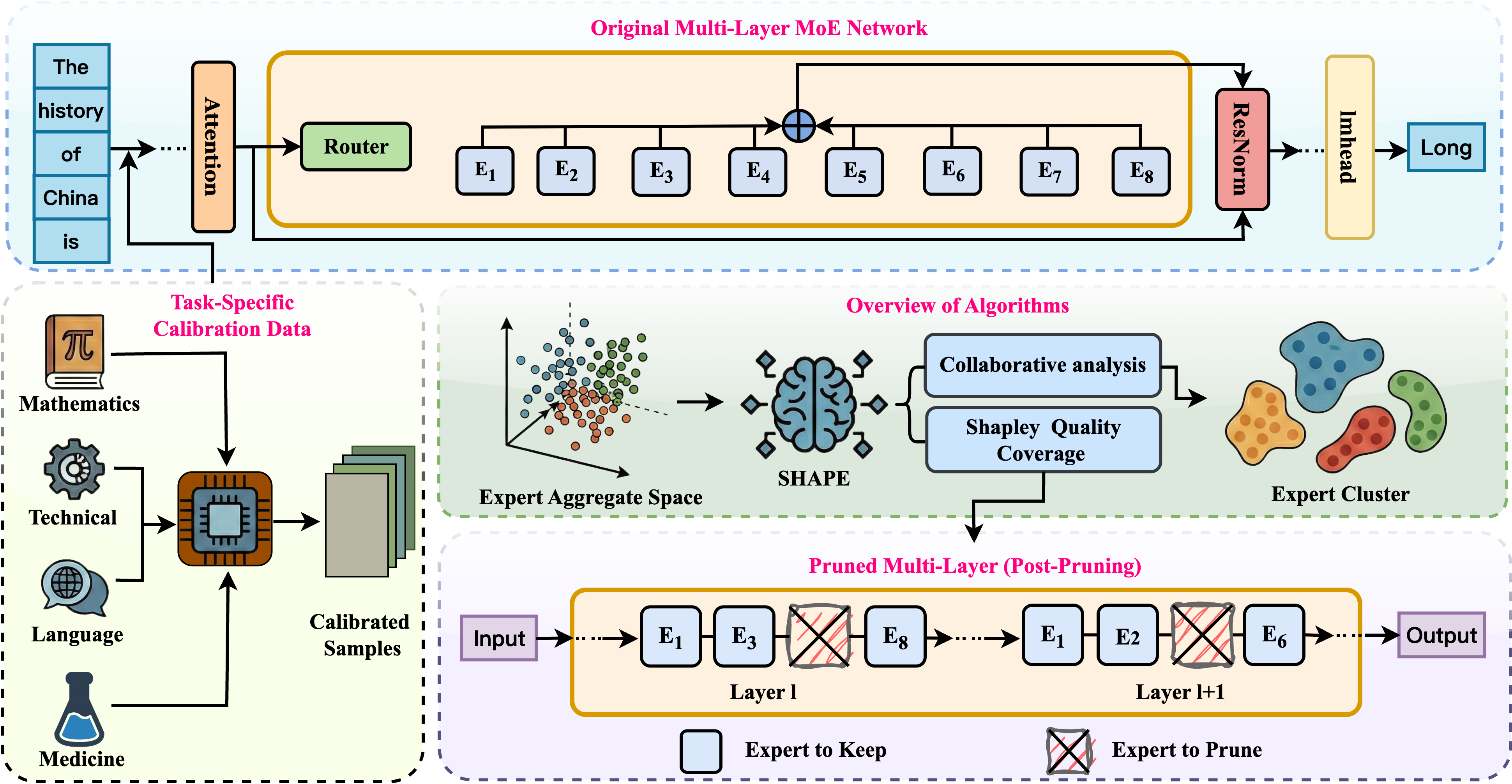}
\caption{Overview of SHAPE. We run an offline calibration pass on a small task-specific dataset to extract routing/co-activation traces and estimate cooperative expert contributions. Based on these estimates, SHAPE performs Shapley-guided quality-coverage expert selection to prune redundant experts, producing a compact MoE model for efficient serving.}
\label{fig:method-overview}
\end{figure*}

The overall workflow of SHAPE is illustrated in Figure~\ref{fig:method-overview}, highlighting the process from expert coalition analysis to Shapley value-based pruning and the post-pruning network.Our contributions are summarized as follows:
\begin{enumerate}
    \item \textbf{Coalitional Expert Valuation.} 
    We introduce a cooperative formulation of MoE pruning based on Shapley values, providing a principled mechanism to attribute expert contributions conditioned on routed coalitions. 
    This work reframes pruning from expert-wise importance estimation to coalition-aware contribution modeling.
    
    \item \textbf{Structure-Preserving Pruning Strategy.} 
    We propose a topology-aware selection mechanism that preserves layer-wise cooperative capacity by retaining experts according to their cumulative Shapley value, rather than enforcing a fixed expert count.
    
    \item \textbf{Empirical Validation of Cooperative Pruning.} 
    Extensive experiments on modern MoE architectures, including Qwen3-MoE, GPT-OSS, and DeepSeek-V2-Lite, demonstrate that preserving expert cooperation is critical for effective compression. 
    SHAPE achieves up to 40\% expert pruning with negligible performance degradation on reasoning-intensive benchmarks.
\end{enumerate}

\section{Related Work}
\label{sec:related}

\subsection{MoE Compression and Expert Pruning}
MoE LLMs achieve \emph{conditional computation} by routing tokens to a sparse subset of experts, but practical serving still requires storing and accessing a large expert pool, resulting in substantial memory footprint and bandwidth pressure~\cite{zoph2022stmoe,mixtral2023}. 
To reduce deployment cost, prior work explores MoE compression via structured pruning~\cite{chowdhury2024provablyeffectivemethodpruning,dong2025domain}, 
expert merging or reparameterization~\cite{zhou2025dernDroppingExpertsRecombiningNeurons,li2025moesvd}, 
quantization~\cite{hu2025moequant,qmoe2023}, 
and knowledge distillation~\cite{xue2022studentknowsexpertsknow,yang2021rethinkingknowledgedistillationperspective}. 
Among them, \emph{expert pruning} is particularly attractive for serving because it directly shrinks the number of experts that must be loaded, cached, and dispatched, yielding immediate reductions in memory and inference overhead.

Recent pruning criteria are often training-free or lightweight and typically rely on \emph{expert-wise} signals. 
SEAP~\cite{liang2025seap} leverages sparse activation statistics to prune experts under fixed budgets, and domain-oriented methods use few-shot demonstrations to select experts that best support target evaluations~\cite{dong2025domain}. 
While effective, these approaches largely score experts in isolation and may overlook interaction effects in top-$k$ routing, where multiple experts are simultaneously activated and aggregated within each MoE layer.

\subsection{Expert Cooperation and Routing-Structured Pruning}
A distinctive characteristic of MoE inference is that the output is formed by aggregating a routed top-$k$ \emph{expert coalition}. 
This coalition structure implies that an expert's utility can be highly context-dependent: removing one expert may alter the behavior of remaining co-activated experts, and the impact may not be predictable from marginal activation frequency alone. 
Motivated by this, recent studies analyze and exploit structured cooperation patterns in MoE.
MoE Pathfinder~\cite{yang2025pathfinder} leverages routing trajectories to guide expert selection, and HSDL~\cite{tang2025collaboration} reveals hidden collaboration patterns via expert co-activation structures that encode functional roles.
These works suggest that \emph{intra-layer} co-activation is a meaningful signal for pruning and that routing distributions contain rich information beyond global expert usage.

Nevertheless, existing routing-aware approaches typically rely on trajectory or co-activation patterns as heuristics, rather than a principled measure of \emph{marginal contribution} under routed coalitions. 
In particular, the within-layer coalition effect—how much an expert helps \emph{in combination} with other simultaneously activated experts under task-driven routing—remains underexplored in pruning objectives. 
Our work targets this gap by valuing experts through their cooperative contributions to routed coalitions, leading to pruning decisions that better preserve task-relevant collaboration.

\subsection{Shapley-Based Cooperative Attribution and Scalable Approximation}
Shapley value~\cite{shapley1953} is a canonical cooperative game-theoretic attribution that quantifies each participant's average marginal contribution across all coalitions. 
It has been applied to neural networks for identifying responsible components and guiding structured pruning, demonstrating advantages in capturing redundancy and interaction effects compared with purely magnitude-based criteria~\cite{ghorbani2020neuronshapleydiscoveringresponsible,ancona2019explainingdeepneuralnetworks}. 
However, exact Shapley computation is NP-hard, making scalable approximation essential for large models.

Practical Shapley estimation commonly relies on Monte Carlo sampling and related estimators~\cite{ghorbani2021datashapleyvaluationefficient,fan2025enhancinginterpretabilityvisionmodels}. 
Despite these advances, most Shapley-based methods focus on neurons/channels or data valuation, and do not directly address conditional computation architectures where coalition formation is governed by routing. 
MoE models are a natural candidate for cooperative attribution because inference is inherently coalition-based (routed top-$k$ experts), yet Shapley value has not been systematically developed as an expert-level pruning criterion under task-driven routing distributions. 
Our work bridges this gap by introducing an efficient Shapley-style approximation tailored to routed expert coalitions for MoE expert selection.

\section{Methodology}
\label{sec:method}

This section introduces \textbf{SHAPE}, a task-conditioned expert pruning framework for Sparse Mixture-of-Experts (SMoE) models. 
Given a pretrained MoE language model and a target downstream task, SHAPE prunes a large fraction of experts \emph{without retraining} while maintaining task performance.

SHAPE is motivated by the coalitional nature of SMoE inference: at each MoE layer, routing activates a top-$k$ set of experts whose outputs are jointly aggregated, so an expert’s value depends on its \emph{cooperative contribution} within routed coalitions rather than in isolation. 
To capture such interaction effects, SHAPE performs (i) \emph{collaborative analysis} on a small task-specific calibration set to estimate per-layer cooperative contributions via an efficient Shapley-style approximation, and (ii) \emph{Shapley Coverage Pruning} to select a compact expert subset that satisfies a global keep-rate budget while preserving layer-wise contribution mass.

\subsection{Task-Driven Calibration and Notation}

We consider a pretrained Sparse Mixture-of-Experts (SMoE) language model with $L$ MoE layers. Each MoE layer $\ell \in \{1,\dots,L\}$ contains a fixed set of experts denoted by $N_\ell = \{E_1, \dots, E_n\}$. Given a token hidden representation $h$, the router at layer $\ell$ computes gating scores over experts and activates a fixed number $k$ of experts according to a top-$k$ routing policy. The output of the MoE layer is obtained by aggregating the outputs of the selected experts. In this work, we do not modify the routing policy or expert parameters, and instead focus on the routing behavior induced by the pretrained model.

Let $\mathcal{T}$ denote the set of downstream tasks. For each task $t \in \mathcal{T}$, we construct a small calibration dataset $\mathcal{D}^{\mathrm{cal}}_t$ using the same prompt format as the evaluation set. The purpose of calibration is not to adapt the model, but to expose the routing policy to inputs drawn from the target task distribution. We perform standard inference with the unpruned model on $\mathcal{D}^{\mathrm{cal}}_t$ and record the routing decisions made at all MoE layers for every processed token.

During inference, each token passing through an MoE layer $\ell$ activates exactly $k$ experts. Collecting these routing traces yields a set of expert combinations that occur under the task-specific input distribution. For a fixed task $t$ and layer $\ell$, we denote by
\[
\mathcal{S}^{t}_{\ell} = \{A_1, A_2, \dots, A_J\}, \quad A_j \subset N_\ell,\ |A_j| = k,
\]
the set of all distinct expert combinations that appear at least once during inference on $\mathcal{D}^{\mathrm{cal}}_t$. Although the number of possible combinations is $\binom{n}{k}$, the number of observed combinations $J$ is typically much smaller, reflecting the structured routing behavior of the model on a specific task.

For each observed expert combination $A_j \in \mathcal{S}^{t}_{\ell}$, we record its activation frequency, defined as the total number of times this exact set of experts is selected by the router across all tokens in $\mathcal{D}^{\mathrm{cal}}_t$. These activation frequencies summarize how expert capacity is allocated by the pretrained model when processing the target task, and serve as empirical statistics characterizing task-conditioned expert cooperation.

Based on these task-specific routing traces, SHAPE proceeds by estimating expert contribution scores for each MoE layer, selecting a subset of experts under a target keep-rate budget, and exporting a pruned model for downstream evaluation. Concretely, for a given task $t$, SHAPE first collects routing traces on $\mathcal{D}^{\mathrm{cal}}_t$, then computes task- and layer-conditioned expert contribution scores, and finally applies a structured selection rule to determine the retained expert sets $\{S^{\ast}_{t,\ell}\}_{\ell=1}^{L}$. Experts outside the selected sets are removed, yielding a pruned model specialized for task $t$. 

Throughout the remainder of this section, we restrict our analysis to the empirical coalition space induced by top-$k$ routing under the task-specific input distribution. We do not assume access to a complete utility function over all expert subsets, nor do we alter the routing mechanism or perform additional forward evaluations beyond standard inference. Instead, the observed expert combinations and their activation frequencies form the sole basis for estimating expert importance and guiding subsequent pruning decisions.We collect routing traces on $\mathcal{D}^{\mathrm{cal}}_t$, estimate per-layer contributions, apply quality-coverage selection, and evaluate the exported pruned model.

\subsection{Coalition-Based Utility and Shapley-Style Expert Valuation}

SHAPE evaluates expert importance from an empirical cooperative-game perspective based solely on routing traces observed during task-specific inference. This design targets deployment-oriented pruning scenarios, where repeated forward evaluation or retraining is impractical. Instead of defining coalition utility via loss re-evaluation, we treat the router’s activation behavior itself as a proxy for task relevance and construct an empirical game over expert combinations actually used by the model.

For a fixed task $t$ and MoE layer $\ell$, let $N_\ell=\{E_1,\dots,E_n\}$ denote the expert set. During inference on the calibration dataset $\mathcal{D}^{\mathrm{cal}}_t$, the router activates exactly $k$ experts per token. We collect the set of all distinct expert combinations that occur at least once:
\[
\mathcal{S}^{t}_{\ell} = \{A_1, A_2, \dots, A_J\}, \quad A_j \subset N_\ell,\ |A_j|=k.
\]
For each observed coalition $A_j$, we record its activation frequency $v(A_j)$, defined as the number of times this expert combination is selected by the router across all tokens in $\mathcal{D}^{\mathrm{cal}}_t$. We interpret $v(A_j)$ as the empirical utility of coalition $A_j$ for task $t$ at layer $\ell$.

For an expert $E_i \in N_\ell$, we focus on the subset of coalitions in which it participates,
\[
\mathcal{S}_{E_i} = \{A_j \in \mathcal{S}^{t}_{\ell} \mid E_i \in A_j\}.
\]
While the frequencies $\{v(A_j)\}_{A_j \in \mathcal{S}_{E_i}}$ reflect how often expert $E_i$ appears in routed coalitions, frequency alone does not capture its marginal contribution, as it ignores whether comparable utility can be achieved without this expert. To estimate the marginal effect of $E_i$ within a coalition $A_j$, we consider the difference
\[
v(A_j) - v(A_j \setminus \{E_i\}),
\]
where $A_j \setminus \{E_i\}$ denotes the $(k-1)$-expert sub-coalition obtained by removing $E_i$.

Such sub-coalitions are not directly observed under top-$k$ routing. We therefore approximate their utility using co-occurrence statistics from $\mathcal{S}^{t}_{\ell}$. Specifically, for each $A_j \in \mathcal{S}_{E_i}$, we count how many observed coalitions contain $A_j \setminus \{E_i\}$ as a subset:
\[
N_{ij} = \big|\{A \in \mathcal{S}^{t}_{\ell} \mid A_j \setminus \{E_i\} \subseteq A\}\big|.
\]
Intuitively, $N_{ij}$ captures the conditional co-occurrence strength of the remaining experts in $A_j$ under the task-specific routing policy. A larger value indicates that these experts frequently co-activate without $E_i$, suggesting a higher degree of substitutability for the removed expert. This frequency-based approximation avoids additional model execution while reflecting empirical expert cooperation patterns.

Directly using $N_{ij}$ as $v(A_j \setminus \{E_i\})$ may violate the monotonicity requirement of cooperative games, namely that a coalition should not obtain lower utility than its sub-coalitions. Accumulated co-occurrence counts may exceed the observed activation frequency of the original coalition, leading to spurious negative marginal contributions. To address this issue, we introduce an expert-specific scaling factor $\alpha_i \in (0,1]$ and define
\[
v(A_j \setminus \{E_i\}) = \alpha_i \cdot N_{ij}.
\]
We select the largest admissible scaling factor that preserves monotonicity across all coalitions involving $E_i$:
\[
\alpha_i = \min_{A_j \in \mathcal{S}_{E_i}} \frac{v(A_j)}{N_{ij}}.
\]
This construction guarantees $v(A_j) \ge \alpha_i N_{ij}$ for all relevant coalitions, enforcing a conservative and order-preserving approximation of sub-coalition utility and preventing overestimation of expert substitutability.

Finally, we define a Shapley-style approximation of expert $E_i$'s contribution as
\[
\phi_{E_i}^{t,\ell}
\;=\;
\frac{1}{|\mathcal{S}_{E_i}|}
\sum_{A_j \in \mathcal{S}_{E_i}}
\frac{1}{|A_j|}
\Big(
v(A_j) - \alpha_i \cdot N_{ij}
\Big).
\]
The factor $1/|A_j|$ reflects equal sharing of coalition utility among participating experts, analogous to the size-based weighting in the classical Shapley value. The resulting score captures both how frequently expert $E_i$ participates in routed coalitions and how indispensable it is within those coalitions.

We emphasize that this formulation does not aim to recover the exact Shapley value of a full cooperative game over all expert subsets. Rather, it provides a principled, task-conditioned approximation tailored to the constrained coalition space induced by top-$k$ routing. In vertical or domain-specialized tasks where expert interactions are highly structured and routing patterns are stable, this empirical valuation offers a reliable signal for expert selection and pruning.

\subsection{Expert Selection via Shapley Quality Coverage}

Given task and layer specific expert contribution scores $\{\phi^{t,\ell}_i\}$, SHAPE performs expert pruning through a quality-coverage-based selection strategy. This stage is agnostic to how the contribution scores are obtained and focuses solely on using them to derive a structured, budget-controlled pruning decision. The key objective is to retain a subset of experts that preserves most of the cooperative value within each MoE layer, while satisfying a global keep-rate constraint.

For each task $t$ and MoE layer $\ell$, we define the total contribution mass of the layer as
\[
T_{t,\ell} = \sum_{i \in N_\ell} \phi^{t,\ell}_{i}.
\]
Negative contributions may arise due to estimation noise or destructive interactions and are excluded to avoid destabilizing the selection process.

Rather than retaining a fixed number of experts per layer, SHAPE enforces a \emph{quality coverage} criterion. Specifically, for a coverage threshold $\alpha \in (0,1]$, we select the smallest subset of experts $S^{\ast}_{t,\ell}(\alpha) \subseteq N_\ell$ such that
\[
\sum_{i \in S^{\ast}_{t,\ell}(\alpha)} \phi^{t,\ell}_{i}
\;\ge\;
\alpha \cdot T_{t,\ell}.
\]
This constraint ensures that an $\alpha$ fraction of the estimated cooperative value in each layer is preserved after pruning. Importantly, coverage is defined in terms of contribution mass rather than expert count, allowing layers with highly concentrated value to retain fewer experts while preventing layers with more diffuse contributions from being overly pruned.

\begin{algorithm}[!htbp]
\small
\caption{SHAPE: Quality-Coverage Expert Selection (Task $t$)}
\label{alg:shape}
\begin{algorithmic}[1]
\REQUIRE Contribution scores $\{\phi^{t,\ell}_{i}\}$; target keep rate $r$; tolerance $\epsilon$
\STATE For each layer $\ell$, set $\phi^{t,\ell}_{i}\leftarrow \max(\phi^{t,\ell}_i,0)$ and $T_{t,\ell}\leftarrow \sum_{i\in N_\ell}\phi^{t,\ell}_{i}$
\STATE Set $\alpha_{\min}\leftarrow 0$, $\alpha_{\max}\leftarrow 1$
\WHILE{$\alpha_{\max}-\alpha_{\min}>\epsilon$}
    \STATE $\alpha\leftarrow(\alpha_{\min}+\alpha_{\max})/2$
    \FOR{$\ell=1$ to $L$}
        \STATE Sort experts in $N_\ell$ by $\phi^{t,\ell}_{i}$ in descending order
        \STATE Select the smallest prefix set $S^\ast_{t,\ell}(\alpha)$ such that
        \STATE \hspace{1em}$\sum_{i\in S^\ast_{t,\ell}(\alpha)}\phi^{t,\ell}_{i}\ge \alpha\cdot T_{t,\ell}$
    \ENDFOR
    \STATE $\mathrm{keep\_rate}(\alpha)\leftarrow\frac{\sum_{\ell=1}^{L}|S^\ast_{t,\ell}(\alpha)|}{\sum_{\ell=1}^{L}|N_\ell|}$
    \IF{$\mathrm{keep\_rate}(\alpha) > r$}
        \STATE $\alpha_{\max}\leftarrow \alpha$ \COMMENT{keep too many experts; decrease $\alpha$}
    \ELSE
        \STATE $\alpha_{\min}\leftarrow \alpha$ \COMMENT{keep too few experts; increase $\alpha$}
    \ENDIF
\ENDWHILE
\STATE $\alpha^\ast \leftarrow \alpha_{\min}$ \COMMENT{largest $\alpha$ with keep rate not exceeding $r$}
\RETURN $\{S^\ast_{t,\ell}(\alpha^\ast)\}_{\ell=1}^{L}$
\end{algorithmic}
\end{algorithm}
\FloatBarrier

The threshold $\alpha$ controls the trade-off between value preservation and pruning aggressiveness: larger $\alpha$ generally yields larger retained sets. SHAPE selects $\alpha$ by bisection to satisfy a target global keep rate $r$. For each candidate $\alpha$, we form $\{S^{\ast}_{t,\ell}(\alpha)\}_{\ell=1}^{L}$ by taking, in every layer, the minimal prefix of experts (ranked by $\phi^{t,\ell}_{i}$) whose cumulative mass reaches $\alpha$-coverage, and then measure the resulting fraction of retained experts across layers. We decrease $\alpha$ if this fraction is above $r$ and increase it otherwise. Algorithm~\ref{alg:shape} summarizes the procedure.

Quality coverage serves as a structure-preserving constraint that mitigates layer collapse, a failure mode in which pruning concentrates retained experts in a small subset of layers. By enforcing value preservation independently at each layer, SHAPE maintains routing stability and functional diversity across depth, which is particularly important for deep MoE Transformers where different layers encode distinct transformations.

Although SHAPE adopts quality coverage as its default selection strategy, the same contribution scores can support alternative pruning policies, such as global ranking across layers or fixed per-layer retention. We include these variants in our ablation study to isolate the effect of the coverage constraint and demonstrate its role in stabilizing performance under aggressive pruning.

\section{Experiments}
\label{sec:exp}

\subsection{Experimental Setup}
\label{subsec:exp-setup}
We evaluate SHAPE on three Mixtral-style sparse MoE LLM backbones (Qwen3-30B-A3B, GPT-OSS-20B, DeepSeek-V2-Lite) across seven benchmarks covering math reasoning (GSM8K, MATH-500), code generation (HumanEval), knowledge QA (GPQA-Diamond), truthfulness (TruthfulQA), sequence labeling (OntoNotes5), and medical QA (MedMCQA).

For each task $t$, we collect routing traces from 25 calibration examples by running the unpruned model and recording per-layer top-$k$ routing decisions and gating weights. We estimate expert importance using a Monte Carlo Shapley approximation and rank experts by the non-negative Shapley mass $\phi^{t,\ell}_{i}$. SHAPE then applies the quality-coverage rule (Algorithm~\ref{alg:shape}) and uses bisection over a global threshold $\alpha$ to meet the target keep rate, performing pruning \emph{without additional training}.

We report task accuracy and an unweighted average over available tasks, under two pruning ratios: 20\% and 40\% .

\begin{table*}[!t]
\centering
\small
\caption{Task accuracy of three MoE LLM backbones under baseline and pruning. ``Avg'' is the arithmetic mean over \emph{available} task scores in the row (missing entries are excluded).}
\label{tab:task-accuracy}
\setlength{\tabcolsep}{3.5pt}
\renewcommand{\arraystretch}{1.05}
\resizebox{\textwidth}{!}{%
\begin{tabular}{llcccccccc}
\toprule
\textbf{Model} & \textbf{Setting} & \textbf{GSM8K} & \textbf{HumanEval} & \textbf{GPQA-D} & \textbf{MATH-500} & \textbf{TruthfulQA} & \textbf{OntoNotes5} & \textbf{MedMCQA} & \textbf{Avg} \\
\midrule
\multirow{3}{*}{Qwen3-30B-A3B} 
& Baseline 
& 96.21 & 95.73 & 58.59 & 96.80 & 77.60 & 87.15 & 68.35 & 82.92 \\
& 20\% pruned 
& 96.74 & 95.12 & 60.10 & 96.80 & 75.64 & 84.42 & 68.18 & 82.43 \\
& 40\% pruned 
& 95.60 & 89.63 & 67.68 & 94.40 & 73.16 & 82.37 & 66.32 & 81.31 \\
\midrule
\multirow{3}{*}{GPT-OSS-20B} 
& Baseline 
& 94.24 & 95.73 & 61.62 & 92.40 & 69.28 & 93.17 & 68.40 & 82.12 \\
& 20\% pruned 
& 95.53 & 93.29 & 62.63 & 95.00 & 69.03 & 93.16 & 68.47 & 82.44 \\
& 40\% pruned 
& 95.07 & 85.41 & 56.57 & 92.20 & 66.10 & 91.25 & 66.54 & 79.02 \\
\midrule
\multirow{3}{*}{DeepSeek-V2-Lite} 
& Baseline 
& 85.52 & 84.15 & 28.28 & 64.20 & 44.43 & 86.74 & 41.26 & 62.08 \\
& 20\% pruned 
& 83.17 & 82.56 & 31.52 & 62.55 & 46.58 & 87.21 & 43.50 & 62.44 \\
& 40\% pruned 
& 81.54 & 75.80 & 26.53 & 59.53 & 42.56 & 86.22 & 39.52 & 58.81 \\
\bottomrule
\end{tabular}
}
\end{table*}

\subsection{Pruning Results Across Backbones}
\label{subsec:multi-backbone}

We evaluate SHAPE across three sparse MoE backbones with different model scales and routing characteristics. Table~\ref{tab:task-accuracy} summarizes task performance under two pruning ratios (20\% and 40\%) together with the unpruned baselines.

Across all backbones, SHAPE preserves strong overall performance under moderate pruning. At a 20\% pruning ratio, the average performance remains comparable to, and in several cases slightly exceeds, that of the unpruned models. This behavior indicates that a non-trivial fraction of experts contributes limited marginal utility under the evaluated tasks, and can be safely removed without harming task accuracy. As pruning becomes more aggressive, performance degradation increases but remains bounded across models, demonstrating that SHAPE can substantially reduce the expert pool while maintaining stable task competence.

The impact of pruning is not uniform across tasks. Reasoning-intensive benchmarks such as GSM8K, HumanEval, and GPQA-Diamond tend to be more sensitive to expert removal, particularly at a 40\% pruning ratio, whereas tasks with stronger local regularities or structured outputs, including OntoNotes5 and MedMCQA, exhibit higher robustness. This contrast suggests that tasks requiring longer dependency chains or compositional reasoning rely more heavily on complementary expert coalitions, making them more vulnerable to the removal of a small number of critical experts.

We further observe clear differences across backbones. Larger and more expressive models, such as Qwen3-30B-A3B and GPT-OSS-20B, demonstrate stronger resilience to pruning, maintaining high accuracy even when 40\% of experts are removed. In contrast, DeepSeek-V2-Lite shows earlier performance degradation, reflecting a smaller redundancy margin and a more compact expert allocation. Nevertheless, even in this setting, SHAPE consistently preserves competitive performance, indicating that coalition-aware expert selection remains effective when model capacity is limited.

Overall, these results highlight that effective MoE pruning depends not only on identifying individually strong experts, but also on preserving cooperative structures that emerge during task execution. By accounting for expert contributions across coalition contexts, SHAPE provides a stable pruning mechanism that generalizes across backbones and task families.

\subsection{Comparison with Baseline Pruning Criteria}
\begin{table*}[!t]
\centering
\scriptsize
\caption{Baseline comparison on Qwen3-30B-A3B (including the unpruned baseline and 20\%/40\% pruning). Best results under the same pruning rate are in bold. ``Avg'' is the arithmetic mean over the reported tasks.}
\label{tab:baseline-comparison-qwen}
\setlength{\tabcolsep}{3.5pt}
\renewcommand{\arraystretch}{1.05}
\resizebox{\textwidth}{!}{%
\begin{tabular}{llcccccccc}
\toprule
\textbf{Rate} & \textbf{Method} & \textbf{GSM8K} & \textbf{HumanEval} & \textbf{GPQA-D} & \textbf{MATH-500} & \textbf{TruthfulQA} & \textbf{OntoNotes5} & \textbf{MedMCQA} & \textbf{Avg} \\
\midrule
\multicolumn{1}{c}{-} & \textbf{Unpruned} 
& 96.21 & 95.73 & 58.59 & 96.80 & 77.60 & 87.15 & 68.35 & 82.92 \\
\midrule
\multirow{6}{*}{20\%} 
& Random 
& 61.24 & 58.36 & 21.42 & 60.27 & 42.31 & 71.16 & 38.18 & 50.42 \\
& Frequency 
& 86.12 & 86.37 & 40.12 & 88.14 & 64.05 & 82.17 & 58.12 & 72.16 \\
& Gating 
& 87.42 & 87.15 & 42.18 & 89.36 & 65.41 & 83.24 & 59.27 & 73.43 \\
& RAEP 
& 95.47 & 94.52 & 57.56 & 95.42 & 76.42 & 86.14 & 67.51 & 81.86 \\
& EASY-EP 
& 95.82 & 95.01 & 58.12 & 95.86 & \textbf{77.14} & \textbf{86.54} & 67.96 & 82.35 \\
& \textbf{SHAPE (Ours)} 
& \textbf{96.74} & \textbf{95.12} & \textbf{60.10} & \textbf{96.80} & 75.64 & 84.42 & \textbf{68.18} & \textbf{82.43} \\
\midrule
\multirow{6}{*}{40\%} 
& Random 
& 35.43 & 28.27 & 8.28 & 32.15 & 21.29 & 60.41 & 15.46 & 28.76 \\
& Frequency 
& 75.14 & 72.11 & 20.16 & 72.29 & 45.18 & 76.28 & 40.17 & 57.33 \\
& Gating 
& 77.19 & 74.32 & 22.54 & 74.28 & 47.21 & 77.31 & 42.15 & 59.29 \\
& RAEP 
& 93.42 & 90.47 & 52.42 & 92.58 & 43.12 & 83.15 & 64.41 & 74.22 \\
& EASY-EP 
& 94.12 & \textbf{91.07} & 53.12 & 93.14 & \textbf{73.28} & \textbf{83.51} & 65.14 & 79.05 \\
& \textbf{SHAPE (Ours)} 
& \textbf{95.60} & 89.63 & \textbf{67.68} & \textbf{94.40} & 73.16 & 82.37 & \textbf{66.32} & \textbf{81.31} \\
\bottomrule
\end{tabular}
}
\end{table*}
\label{subsec:baseline-compare}
We compare SHAPE with representative baseline pruning criteria across all evaluated backbones. While similar trends are observed on GPT-OSS-20B and DeepSeek-V2-Lite, we use Qwen3-30B-A3B as a representative example to present detailed task-level comparisons. This choice is made for clarity of exposition rather than model specificity.

Overall, simple pruning heuristics based on random selection, activation frequency, or router gating scores exhibit clear limitations. Although frequency- and gating-based methods outperform random pruning, they still incur noticeable performance degradation even at moderate pruning ratios. In particular, reasoning-intensive tasks such as GSM8K, HumanEval, and GPQA-Diamond are highly sensitive to these heuristics, suggesting that routing statistics alone are insufficient to capture task-relevant expert utility. Similar relative gaps between naive heuristics and task-aware methods are consistently observed across backbones.

Stronger baselines, including RAEP~\cite{lasby2025reap} and EASY-EP~\cite{dong2025domain}, substantially improve pruning stability by incorporating output-aware or task-aligned signals. At a 20\% pruning ratio, these methods largely preserve baseline performance on most tasks, indicating that expert selection guided by task feedback is crucial for effective pruning. However, as pruning becomes more aggressive, their robustness degrades across multiple models. Performance drops become especially pronounced on tasks that rely on compositional reasoning or long dependency chains, implying that expert-wise importance estimation alone cannot fully preserve the underlying computation structure.

Across all pruning ratios, SHAPE consistently achieves the strongest or near-strongest performance. Its advantage becomes most evident at 40\% pruning, where maintaining a small number of critical experts is decisive. On Qwen3-30B-A3B, SHAPE retains substantially higher accuracy on GPQA-Diamond and GSM8K compared to all baselines, and the same qualitative behavior is observed on the other backbones. This consistency across models suggests that SHAPE captures backbone-agnostic properties of expert cooperation rather than exploiting architecture-specific artifacts.

Taken together, these results indicate that the primary limitation of existing pruning criteria lies in treating experts as largely independent units. In sparse MoE inference, however, effective computation emerges from structured cooperation among a small subset of experts. By explicitly valuing experts through their marginal contributions across coalition contexts, SHAPE more reliably preserves these cooperative structures, leading to improved robustness under aggressive pruning.

\subsection{Ablation on Pruning Strategies}
\label{subsec:strategy-ablation}

We analyze the effect of different pruning strategies under the same expert budget to isolate the role of the quality-coverage constraint in SHAPE. Table~\ref{tab:strategy-ablation} reports average task performance at 20\% and 40\% pruning across three backbones, comparing SHAPE with two simplified variants: global pruning based on a single Shapley ranking and layer-wise pruning that selects top experts independently within each layer.

Across backbones and pruning ratios, SHAPE consistently achieves the most stable performance. In contrast, global pruning exhibits increasing instability as pruning becomes more aggressive. By selecting experts solely according to a global ranking, this strategy can over-prune layers that contain long-tail specialists while over-allocating capacity to others, leading to structural imbalance in the MoE topology. This effect becomes more pronounced at 40\% pruning, where removing a small number of layer-specific experts can disproportionately degrade end-to-end performance.

Layer-wise pruning mitigates extreme layer collapse by enforcing a fixed budget per layer, but still underperforms SHAPE in most settings. While selecting experts independently within each layer preserves locally strong specialists, it fails to account for how capacity should be distributed across layers according to task demands. As a result, layers that require higher cooperative capacity for a given task may remain under-provisioned, limiting the model’s ability to form effective expert coalitions.
\begin{table}[!htbp]
\centering
\small
\caption{Ablation of pruning strategies using the Avg metric at 20\% and 40\% pruning.}
\label{tab:strategy-ablation}
\begin{tabular}{llccc}
\toprule
\textbf{Strategy} & \textbf{Rate} & \textbf{Qwen3-Moe} & \textbf{GPT-OSS} & \textbf{DeepSeekV2} \\
\midrule
\multirow{2}{*}{SHAPE} 
& 20\% & 82.43 & 82.44 & 62.44 \\
& 40\% & 81.31 & 77.53 & 58.81 \\
\midrule
\multirow{2}{*}{Global} 
& 20\% & 82.21 & 81.62 & 61.53 \\
& 40\% & 80.49 & 76.92 & 57.64 \\
\midrule
\multirow{2}{*}{Layer-wise} 
& 20\% & 81.53 & 80.85 & 60.86 \\
& 40\% & 79.58 & 75.83 & 56.84 \\
\bottomrule
\end{tabular}
\end{table}

By contrast, SHAPE explicitly enforces a minimum retained Shapley mass per layer through the quality-coverage constraint. This mechanism prevents both global imbalance and rigid per-layer allocation, encouraging a balanced expert set that maintains cooperative structures across the network. The resulting pruned models exhibit consistently higher average performance, particularly under aggressive pruning, demonstrating that coverage-aware expert selection is crucial for preserving MoE functionality under constrained expert budgets.

\subsection{Results and Analysis}
\label{subsec:results-analysis}

\begin{figure}[!htbp]
\centering
\includegraphics[width=\columnwidth]{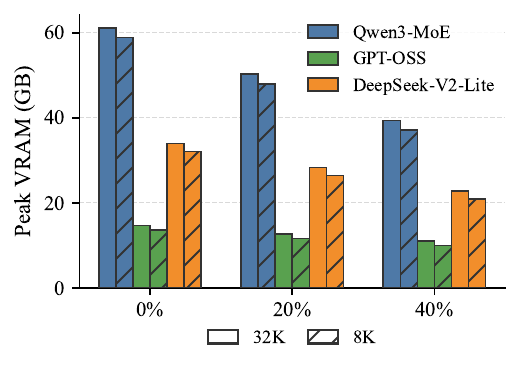}
\caption{Peak GPU memory (VRAM) comparison under 32K context and batch size 1. Although MoE activates only a few experts per token, serving the full expert pool still incurs substantial memory overhead; expert pruning directly reduces the served footprint.}
\label{fig:memory-peak}
\end{figure}

We further analyze the deployment implications of expert pruning by examining peak GPU memory usage. Fig.~\ref{fig:memory-peak} reports the peak VRAM consumption under a 32K context length and batch size 1, comparing the unpruned model with pruned variants at different pruning ratios.

Although sparse MoE models activate only a small subset of experts per token, serving the full expert pool still incurs substantial memory overhead due to parameter storage and runtime bookkeeping. As shown in Fig.~\ref{fig:memory-peak}, pruning leads to a clear and consistent reduction in peak memory usage, and the savings grow with more aggressive expert removal. This indicates that memory consumption in practical MoE serving is dominated by the size of the resident expert pool rather than by the number of activated experts during inference.

Importantly, the observed memory reduction is achieved without additional training or architectural modification, highlighting the practicality of pruning as a deployment-oriented optimization. Combined with the accuracy results in the previous sections, these findings suggest that SHAPE enables a favorable trade-off between model capacity and resource footprint, making large sparse MoE models more amenable to deployment under constrained GPU memory budgets.

\section{CONCLUSION}
\label{sec:concl}
We proposed \textbf{SHAPE}, a task-driven expert pruning framework for sparse MoE language models that respects the coalitional nature of routed top-$k$ inference by valuing experts through cooperative contributions estimated from task-specific routing traces, and selecting experts with a layer-aware quality-coverage rule that matches a target keep rate via bisection. Across multiple modern MoE backbones and diverse benchmarks, SHAPE consistently preserves accuracy better than global and layer-wise pruning variants under 20\% and 40\% pruning, while reducing the served expert footprint and peak GPU memory, making MoE deployment more practical in resource-constrained settings.

\bibliographystyle{IEEEtran}
\bibliography{refs}
\end{document}